\pdfoutput=1
\documentclass[11pt]{article}
\usepackage{coling2020}
\usepackage{times}
\usepackage{url}
\usepackage{latexsym}
\usepackage{adjustbox}
\usepackage{booktabs}
\usepackage{fixltx2e}
\usepackage{multirow}
\usepackage{sidecap}
\usepackage{graphics}
\usepackage{graphicx} 
\usepackage[subrefformat=parens]{subcaption}
\usepackage{xcolor}
\usepackage{caption}
\usepackage{amsmath}
\usepackage{caption,stackengine}
\usepackage{parselines} 
\usepackage{xspace} 
\usepackage[symbol]{footmisc}

\colingfinalcopy 

\title{I Know What You Asked: \\ Graph Path Learning using AMR for Commonsense Reasoning}

\author{Jungwoo Lim\footnote[1]{}, Dongsuk Oh\footnote[1]{}, Yoonna Jang, Kisu Yang, Heuiseok Lim\footnote[2]{}   \\
Computer Science and Engineering, Korea University \\
Republic of Korea \\
{\tt \{wjddn803,inow3555,morelychee,willow4,limhseok\}}@korea.ac.kr \\}
\date{}

\begin{document}

\maketitle

\footnotetext{\footnote[1]{} Equal contribution}
\footnotetext{\footnote[2]{} Corresponding author}

\begin{abstract}
\texttt{CommonsenseQA} is a task in which a correct answer is predicted through commonsense reasoning with pre-defined knowledge. Most previous works have aimed to improve the performance with distributed representation without considering the process of predicting the answer from the semantic representation of the question.
To shed light upon the semantic interpretation of the question, we propose an AMR-ConceptNet-Pruned (ACP) graph.
The ACP graph is pruned from a full integrated graph encompassing Abstract Meaning Representation (AMR) graph generated from input questions and an external commonsense knowledge graph, ConceptNet (CN). Then the ACP graph is exploited to interpret the reasoning path as well as to predict the correct answer on the \texttt{CommonsenseQA} task.
This paper presents the manner in which the commonsense reasoning process can be interpreted with the relations and concepts provided by the ACP graph. Moreover, ACP-based models are shown to outperform the baselines.
\end{abstract}

\section{Introduction}
\label{intro}
\blfootnote{
    %
    %
    \hspace{-0.65cm}  
    This work is licensed under a Creative Commons Attribution 4.0 International Licence. Licence details: \url{http://creativecommons.org/licenses/by/4.0/}.
    %
    %
    %
    %
}

\begin{figure}
\begin{subfigure}[b]{.31\linewidth}

\centering
\includegraphics[width=\linewidth]{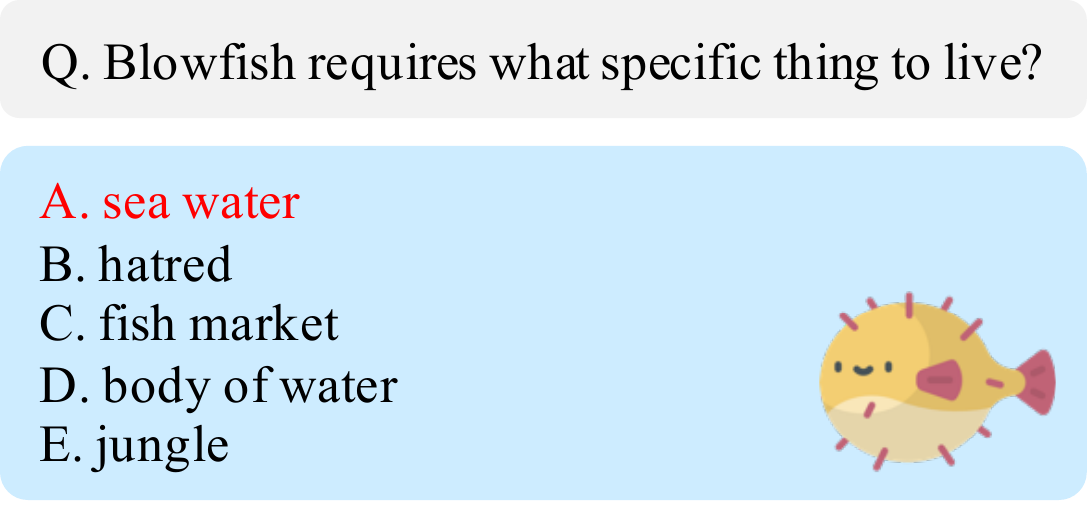}
\caption{Example of \texttt{CommonsenseQA} dataset}
\label{subfig:examplea}
\end{subfigure}
\hfill
\begin{subfigure}[b]{.31\linewidth}
\centering
\includegraphics[width=\linewidth]{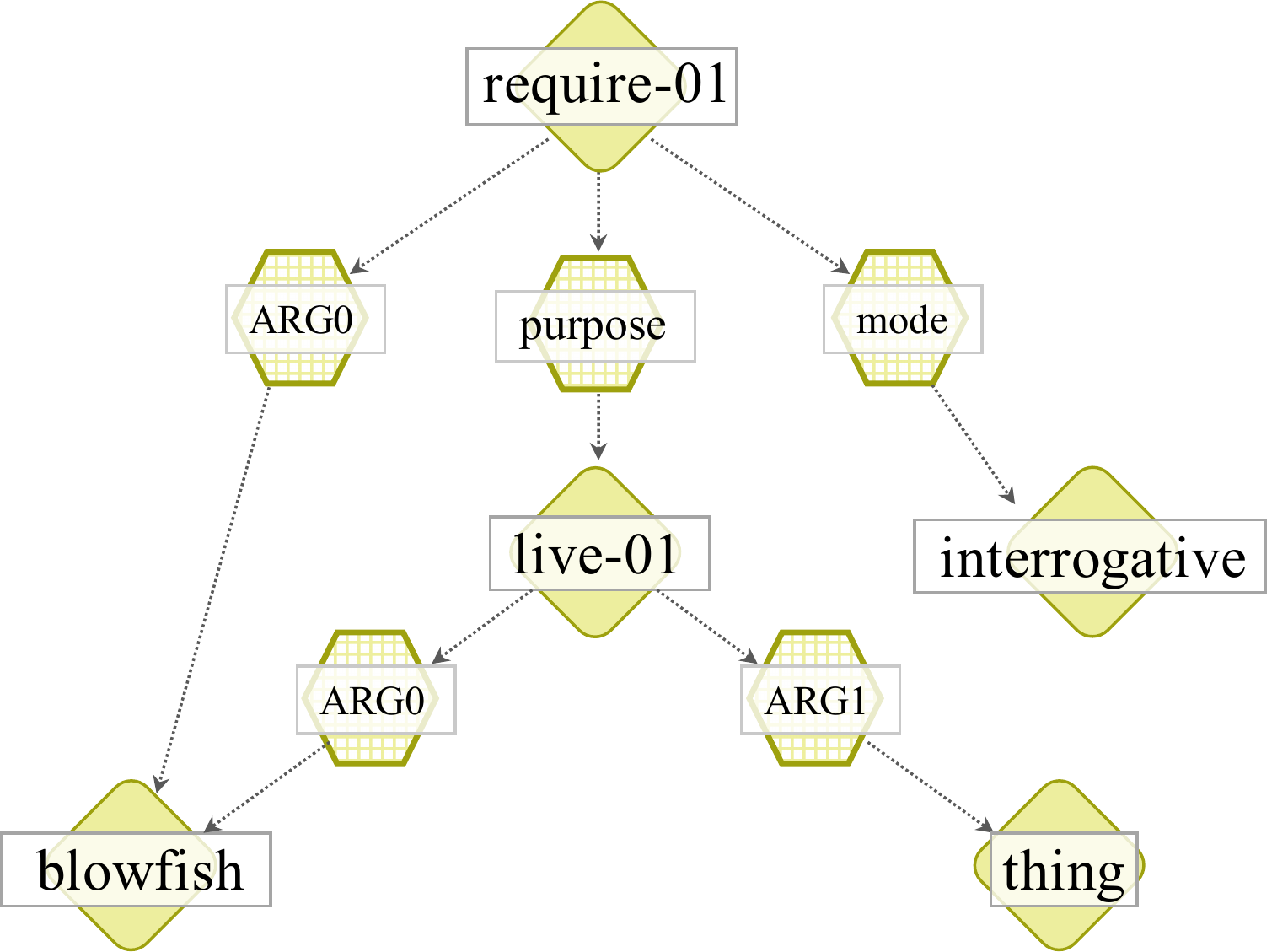}
\caption{Example of AMR graph of question}
\label{subfig:exampleb}
\end{subfigure}
\hfill
\begin{subfigure}[b]{.31\linewidth}
\centering
\includegraphics[width=\linewidth]{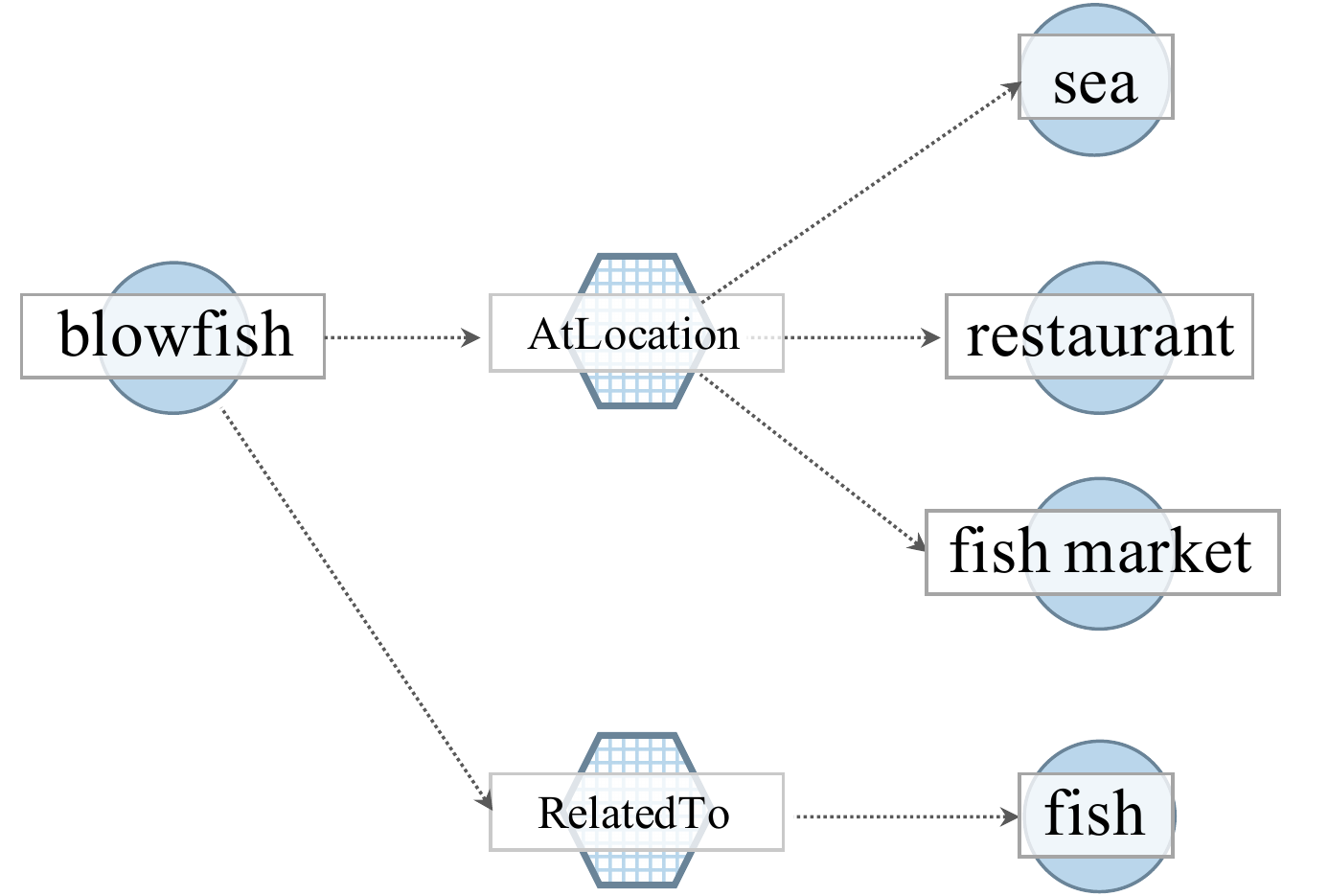}
\caption{Example of ConceptNet subgraph of question}
\label{subfig:examplec}
\end{subfigure}%
\caption{Example of AMR and ConceptNet graphs of question.}
\label{fig:example}
\end{figure}

Commonsense is the knowledge shared by the majority of people in society and acquired naturally in everyday life. Commonsense reasoning is the process of logical inference by using commonsense information. Commonsense to answer the questions that is ``\texttt{Blowfish requires what specific thing to live?}'' in Figure \ref{fig:example} is depicted as: ``\texttt{Blowfish is fish}'', ``\texttt{Fish lives in the water}'', and ``\texttt{Water includes seas and rivers}.'' An enormous amount of pre-defined commonsense knowledge is available and people can make inferences using this commonsense such as in the following example: ``\texttt{Blowfish is fish.}'' $\rightarrow$ ``\texttt{Fish lives in the water.}'' $\rightarrow$ ``\texttt{Water includes seas and rivers.}'' $\Rightarrow$ ``\texttt{Blowfish lives in the sea.}'' This chain of commonsense reasoning is naturally deduced by humans without substantial difficulty. Whereas people acquire commonsense in their lives, machines cannot learn this knowledge without any assistance. A large amount of external knowledge and several reasoning steps are required for machines to learn commonsense. In recent years, various datasets ~\cite{zellers2018swag,sap2019socialiqa,zellers2019recognition} have been constructed to enable machines to reason commonsense.

\texttt{CommonsenseQA} \cite{talmor2019commonsenseqa} is one of the most widely researched datasets and is presented in Figure \ref{fig:example} \subref{subfig:examplea}. The studies of commonsense reasoning based on this dataset can be categorized into two mainstream approaches. The first approach uses pre-trained language models with distributed representations, which exhibit high performances on most Natural Language Processing (NLP) tasks. However, despite their high performance, these models must be trained with an excessive number of parameters and cannot explain the process of commonsense reasoning. The second approach is reasoning with a commonsense knowledge graph. The generally used commonsense knowledge graph is ConceptNet 5.5 \cite{speer2017conceptnet}, which includes parsed representation from Open Mind Commonsense (OMCS) and other different language sources such as WordNet \cite{bond2013linking} or DBPedia \cite{auer2007dbpedia}. In this approach, the subgraph of  ConceptNet corresponding to the questions are transformed into node embeddings by the graph encoder. The candidate with the highest attention score is selected as an answer that is computed between the node embeddings and the word vectors from the language models. To learn the commonsense knowledge that is not observed or understood by the language models, relations from ConceptNet serve as a critical role in this method.
The performance is improved by utilizing the relations that are not represented in the text; however, the interpretation of the question is still not enough. 

Unlike \texttt{CommonsenseQA}, the most commonly used method of solving this problem is Knowledge-Based Question-Answering (KBQA) \cite{berant2013semantic,yih2014semantic,yih2015semantic} employing semantic representations. As this method infers the answer with the logical structure of the question using the knowledge base, the question-answering process can be explained in a logical form. In our work, Abstract Meaning Representation (AMR) \cite{banarescu2013abstract}, which is one of the logical structure, is used to understand the overall reasoning process, from the question to the answer.
AMR is a graph for meaning representation that symbolizes the meaning of sentences. AMR illustrates ``who is doing what to whom'' that is implied in a sentence with a graph. 
The components of these graphs are not the words, but rather the concepts and their relations. Each concept denotes an event or an entity, and each relation represents the semantic role of the concepts. 

In this paper, we enable the language models to exploit the AMR graph to understand the logical structure of sentences. However, it is difficult to infer commonsense information with only an AMR graph, owing to its deficiency of commonsense knowledge of the given sentence. For example, in Figure \ref{fig:example} \subref{subfig:exampleb}, the AMR graph indicates the path of the logical structure of the sentence ``\texttt{What the blowfish requires to live?}'' (\texttt{require-01} $\rightarrow$ \texttt{purpose} $\rightarrow$ \texttt{live-01} $\rightarrow$ \texttt{ARG0} $\rightarrow$ \texttt{blowfish}); in other words, these paths from the single AMR graph lack the proficient information to predict the right answer. Therefore, for commonsense reasoning, dynamic interactions between the AMR graph and ConceptNet are inevitable to reach the correct answer.

Thus, we propose a new compact AMR graph expanded with the ConceptNet's commonsense relations with pruning, and it is called ACP graph.
The proposed method can interpret the path from the question to the answer by performing commonsense reasoning within the connected graph, such as
``\texttt{Blowfish needs the sea to live.}'' (\texttt{require-01} $\rightarrow$ \texttt{purpose} $\rightarrow$ \texttt{live-01} $\rightarrow$ \texttt{ARG0} $\rightarrow$ \texttt{blowfish} $\rightarrow$ \texttt{AtLocation} $\rightarrow$ \texttt{sea}).

\vspace{5mm}

The contributions of our study are as follows.

\begin{itemize}

\item We introduce a new graph structure the ACP graph, which is pruned from a full integrated graph encompassing Abstract Meaning Representation (AMR) graph generated from input questions and an external commonsense knowledge graph, ConceptNet (CN) for commonsense reasoning. This structure is represented in a Levi graph \cite{gross2013handbook} form to enable relation interpretations.

\item We propose a graph-path reasoning framework, using which it is possible to explain the path from the question to the answer in a logical manner based on commonsense reasoning.

\item Our path reasoning method exhibits a performance improvement over previous models.
\end{itemize}

\vspace{5mm}

The remainder of this paper is organized as follows. In Section 2, we present the entire process of our method in detail. The experimental setup and results are explained in Section 3. A discussion of the proposed model is provided in Section 4, and Section 5 presents the conclusions. Appendix A provides related works including ConceptNet, previous works on commonsense reasoning, and AMR.

\section{Proposed Method}
\label{model}

\begin{figure}[t]
	\centering
	\includegraphics[width=0.9\textwidth]{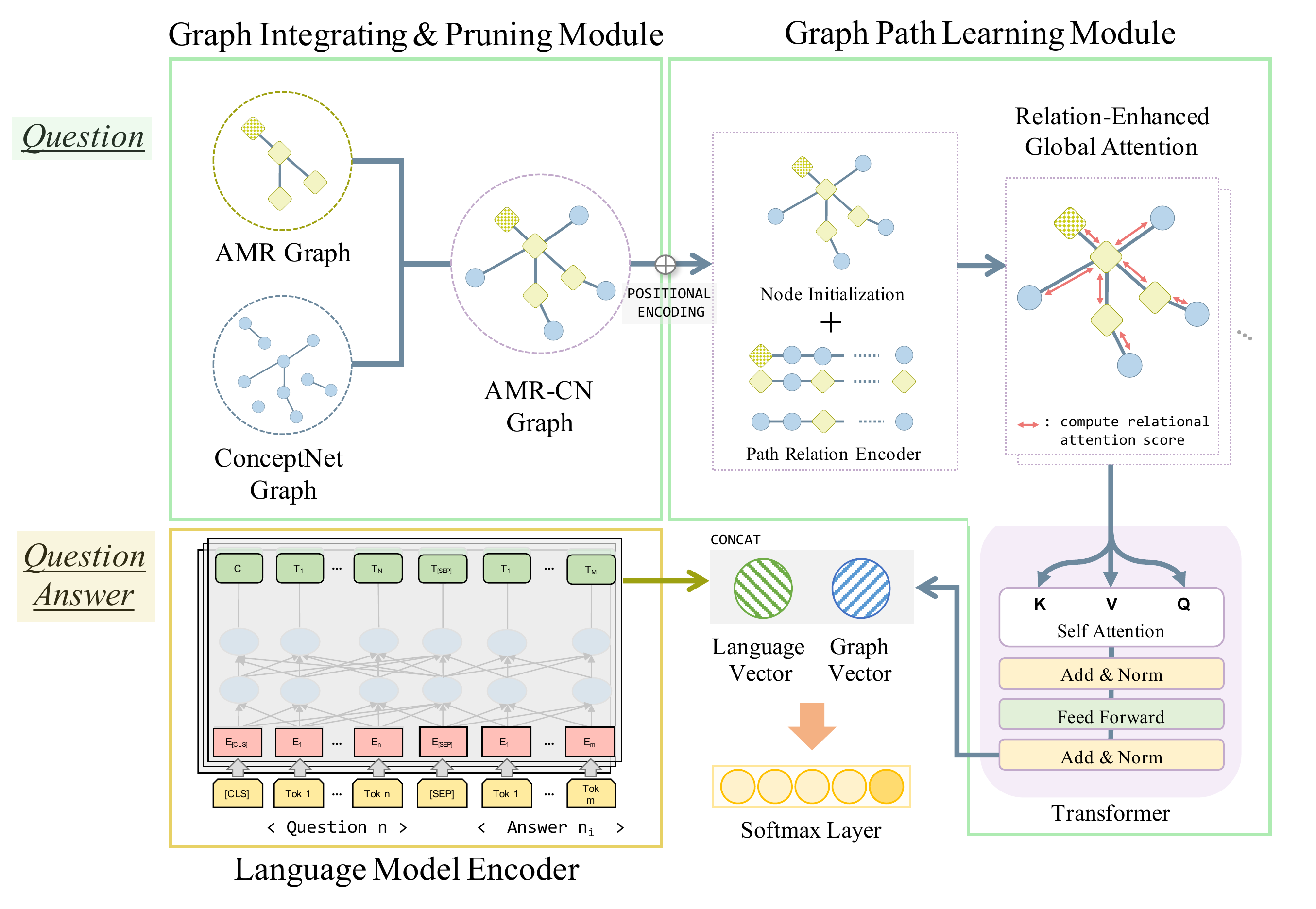}
	\caption{Overview of proposed method.}  
	\vspace{-3mm}
\label{fig:model}
\end{figure}

We propose a commonsense reasoning framework that uses a commonsense knowledge base on the basis of the AMR logic structure.
Our framework consists of the AMR graph integrating and pruning module, language model encoder, and graph path learning module\footnote[4]{Code  available at \url{https://github.com/dlawjddn803}}. As illustrated in Figure \ref{fig:model}, we first generate the AMR graph from every question in the \texttt{CommonsenseQA} dataset and integrate all the nodes of AMR with ConceptNet graphs. 

As this AMR-ConceptNet full graph also includes some irrelevant relations to the question, interpreting questions can be guided in the wrong way. For this reason, we suggest a new method, the ACP graph, pruned according to the relation type.
Thereafter, the graph path learning module takes the pruned graph as an input and computes the attention score of each path by using the Graph Transformer \cite{cai2019graph:2019} which results in the whole graph vector. The graph vector is finally fed into the Transformer \cite{vaswani2017attention} to model the interactions between the AMR and ConceptNet graph and transforms to the final graph representation. Meanwhile, the question and candidate answer from the dataset are passed through the language model encoder, producing the language vector. The concatenation of the language and graph vectors turns out to the final representation that is used to predict the correct answer.


In contrast to other models mentioned in Talmor \shortcite{talmor2019commonsenseqa}, which cannot provide interpretable reasons for predicting the correct answer from the question, our proposed method produces the reasoning paths that make the model transparent and interpretable. That is, the reasoning paths that have high attention weights from the graph encoder possess potentially accurate information for reasoning. These reasoning paths are depicted in Figure \ref{fig:case-amrcn}.

\subsection{Graph Integrating and Pruning}
\begin{figure}
\begin{subfigure}[b]{.52\linewidth}
\centering
\includegraphics[width=\linewidth]{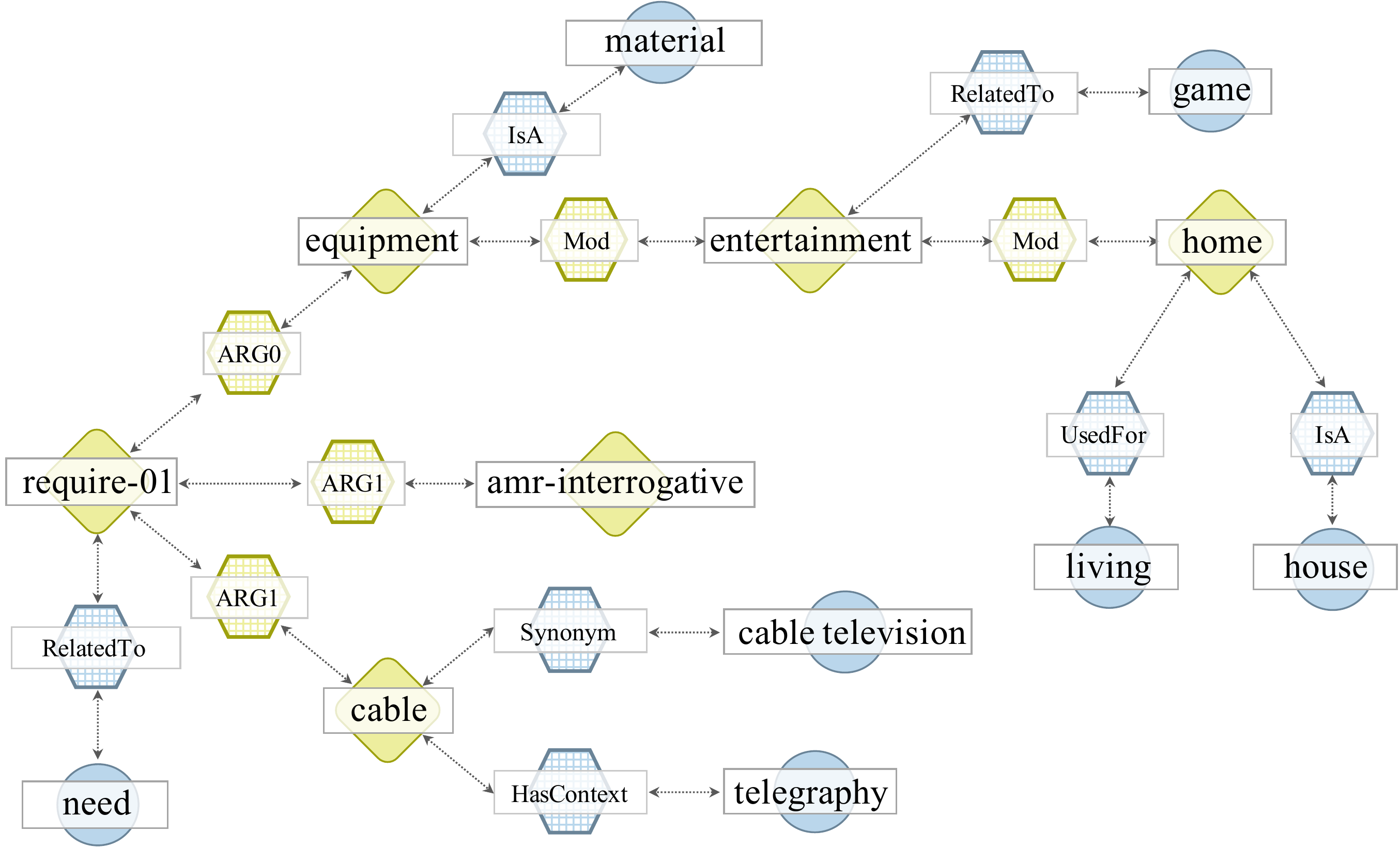}
\caption{AMR-CN graph -- full}
\label{fig:amr_cn_a}
\end{subfigure}
\hfill
\begin{subfigure}[b]{.46\linewidth}
\centering
\includegraphics[width=\linewidth]{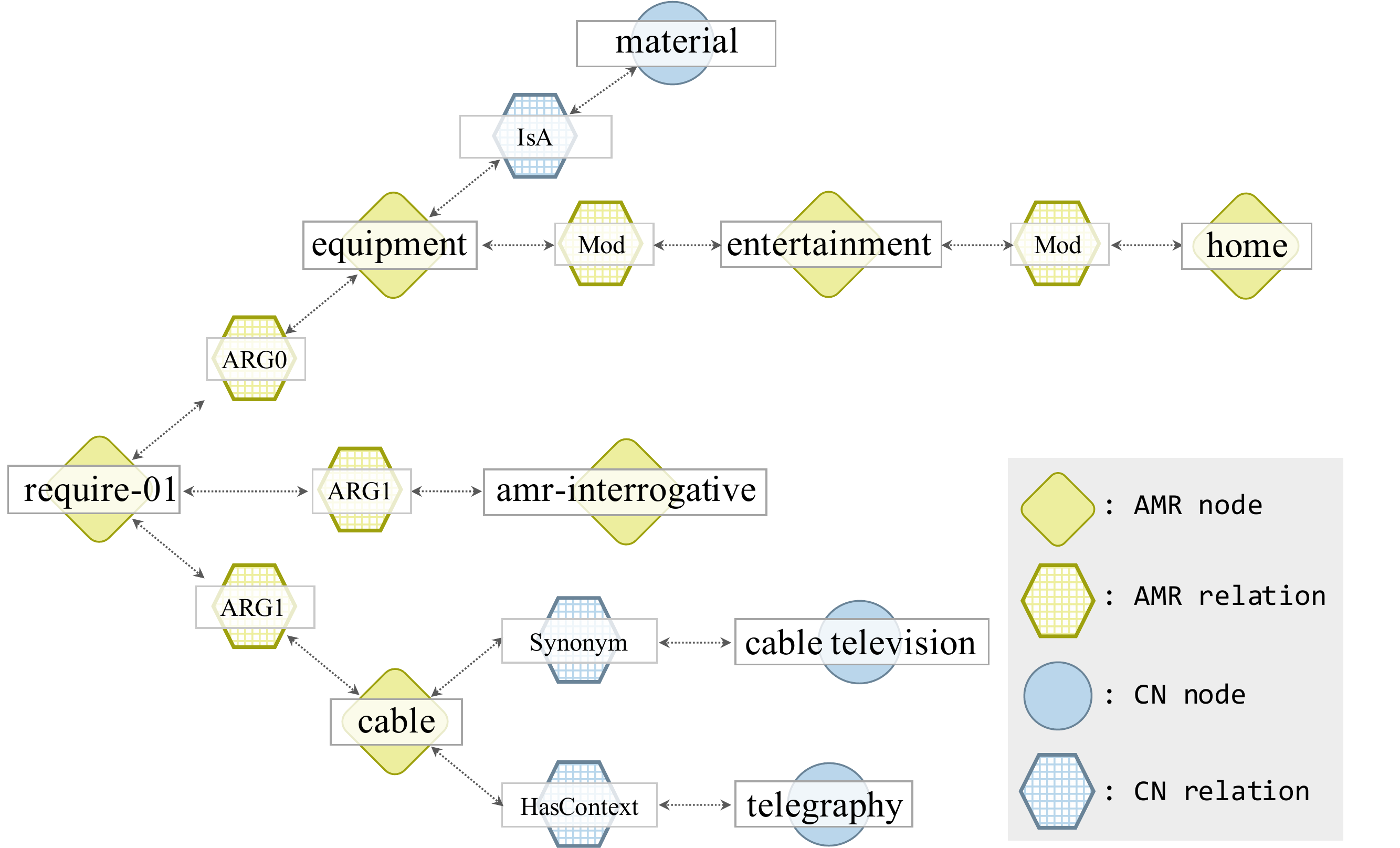}
\caption{AMR-CN graph -- pruned}
\label{fig:amr_cn_b}
\end{subfigure}%
\caption{Two methods of integrating AMR graph with ConceptNet graph. Method \subref{fig:amr_cn_a} incorporates all of the nodes from the AMR graph with the ConceptNet graph. In contrast, method \subref{fig:amr_cn_b} removes the ConceptNet nodes that is not connected to the AMR nodes which have \texttt{ARG0} and \texttt{ARG1} relations. For instance, we generate an AMR graph from the sentence ``\texttt{What home entertainment equipment requires cable?}", and we find every ConceptNet relation that includes all the AMR nodes such as \texttt{(cable-HasContext-telegraphy)}, \texttt{(require01-RelatedTo-need)}, and \texttt{(home-UsedFor-living)}. Then remove the ConceptNet nodes that are not connected to the AMR nodes which have \texttt{ARG0} or \texttt{ARG1} relations. In this example, the deleted nodes are \texttt{living}, \texttt{house}, and \texttt{game}. In addition, as the \texttt{require-01} is frame node which does not need to expand, \texttt{need} is also removed from the graph. 
}

\label{fig:amr_cn}
\end{figure}

As each word plays a certain role as a predicate or an argument in a sentence, the concepts of the AMR graph also carry semantic meanings in the graph structure. Hence, the AMR graph is capable of interpreting the questions as paths, semantically . Owing to these advantages of the graph structure and preserved semantic interpretation, we use the AMR graph for extracting commonsense knowledge graph. To generate an AMR graph from the raw text, we use the pre-trained model of Zhang \shortcite{zhang2019amr:2019}, which is an attention-based model that treats AMR parsing as sequence-to-graph transduction. Though most of the AMR graphs generated from the model properly, they might have some inevitable errors in the type of relations or concept.
\vspace{-6mm}

\begin{center}
\begin{table}[b]
{
\centering
\scalebox{0.80}
	{

	\begin{tabular}{c|rrrr}
		\toprule   
		 \textbf{Relation} & \textbf{Ntrain} & \textbf{Ndev}  & \textbf{Ntest}  \\
		\midrule\midrule
			\textbf{ARG0} & \textbf{17300 (22.70\%)} & \textbf{2547 (22.73\%)} & \textbf{2477 (23.09\%)}  \\
		    \textbf{ARG1} & \textbf{24673 (32.38\%)} & \textbf{3566 (31.83\%)} & \textbf{3521 (32.82\%)}  \\
		    ARG2 & 6001 (7.88\%) & 864 (7.71\%) & 829 (7.73\%)  \\
		    ARG3 & 286 (0.38\%) & 37 (0.33\%) & 51 (0.48\%)  \\
		    ARG4 & 587 (0.77\%) & 92 (0.82\%) & 59 (0.55\%)  \\
		\midrule
		    Total relations & 76203 & 11204 & 10727 \\
		\bottomrule

	\end{tabular}

	}
	\caption{Statistics of core roles in \texttt{CommonsenseQA} AMR graph. We split the given training set into the new training and test sets randomly to conduct diverse experiments with efficiency. The new training, development, and test sets included 8,500, 1,221, and 1,241 examples, respectively. \vspace{12pt}}
	\label{table:corerole}

}
\end{table}
\end{center}

We suggest effective AMR expansion and pruning rules for commonsense reasoning. We expand the AMR graph on all nodes with the ConceptNet as illustrated in Figure \ref{fig:amr_cn} \subref{fig:amr_cn_a}, and prune the nodes that have edges known as \texttt{ARG0} and \texttt{ARG1} with ConceptNet. Considering that \texttt{ARG0} and \texttt{ARG1} are the top two frequent relations among any other relations as shown in Table \ref{table:corerole}, we prune the full AMR-CN graph into a more compact graph that only contains \texttt{ARG0} and \texttt{ARG1} relations, which is called ACP graph. This procedure prevents the graph from discovering a tremendous number of paths iteratively. As described in Appendix A, since the frame node is defined as a central point in the AMR graph like \texttt{require-01} in Figure \ref{fig:amr_cn}, combining other ConceptNet relations with the root node may distract the process of path reasoning. Also, the frame node's specific meaning additionally annotated by the number like ``\texttt{-01}'' at the end of the word is different from the meaning in ConceptNet's node even though it has identical letters. For example, the specific meaning of the frame node ``\texttt{play-11}'' is ``play/perform music'' defined in Propbank frameset while ConceptNet's node ``\texttt{play}'' includes more diverse meanings such as ``engage in an activity like game''or ``bet or wager''. Therefore, we remove the ConceptNet relations and nodes connected to the frame node. The proposed method is depicted in Figure \ref{fig:amr_cn} \subref{fig:amr_cn_b}.

\vspace{3mm}

The graph G = (V, E) expresses fixed set of nodes V, and relation edges E. Following this notation, the ACP graph is defined as follows: 
\begin{equation}
    \mathcal{G_\mathnormal{ACP}}=(\{\mathcal{V_{\mathnormal{amr}}} \cup {\mathcal{V}_\mathnormal{cn}^\mathit{\tiny amr^{arg}}} \},\{\mathcal{E_{\mathnormal{amr}}} \cup {\mathcal{E}_\mathnormal{cn}^\mathit{\tiny amr^{arg}}} \})
\label{acpgraph}
\end{equation}

The ACP graph expressed in equation (\ref{acpgraph}) is the union set of the AMR and the subgraph of ConceptNet that contains AMR concepts that are connected to \texttt{ARG0} and \texttt{ARG1}, respectively. The AMR graph is denoted as $\mathcal{G_\mathnormal{AMR}}= \{\mathcal{V_{\mathnormal{amr}}},\mathcal{E_{\mathnormal{amr}}}\}$. The subgraph of ConceptNet matched with the concepts that are connected to \texttt{ARG0} and \texttt{ARG1} is defined as $\mathcal{G_\mathnormal{CN}^\mathnormal{AMR^{arg}}}= \{\mathcal{V_{\mathnormal{cn}}^\mathnormal{amr^{arg}}},\mathcal{E_{\mathnormal{cn}}^\mathnormal{amr^{arg}}}\}$

\subsection{Language Encoder and Graph Path Learning Module}

The proposed method performs commonsense reasoning over the ACP graph and predicts the correct answer with the corresponding inference. Our model receives two types of inputs, which are text and graph, and converts semantic representation to distributed representation. To encode the text input into the distributed representation, the language encoder which is the pre-trained language model with a massive amount of corpus takes an input that is formalized as ``\texttt{[CLS]+Question+[SEP]+candidate answer}.'' 
Given the ACP graph from the graph integrating and pruning module, the graph path learning module initializes the concept node vectors as the sum of the concept embedding using GloVe \cite{pennington2014glove} and absolute position embedding. Inspired by the works of Cai \shortcite{cai2019graph:2019}, we modify the graph transformer to make the model reason over the relation paths of the ACP graph. To let the model recognize the explicit graph paths, we first encode the relation between two concepts into a distributed representation using the relation encoder. The relation encoder identifies the shortest path between two concepts and represents the sequence as a relation vector by employing recurrent neural networks with a 
Gated Recurrent Unit (GRU) \cite{cho2014learning}. The equation for the represented relation is expressed as follows:

\begin{equation}
    \overrightarrow{p_{t}} = \mathnormal{\mathrm{GRU}_{f}( \overrightarrow{p_{t-1}},sp_{t})},   
    \overleftarrow{p_{t}} = 
    \mathnormal{\mathrm{GRU}_{g}( \overleftarrow{p_{t+1}},sp_{t})}
\label{eq:relencoder}
\end{equation}

where $\mathnormal{sp_{t}}$ indicates the shortest path of the relation between two nodes. The final relation encoding $\mathnormal{r_{ij}}$ between concepts $\mathnormal{i}$ and $\mathnormal{j}$ is the concatenation of the final hidden states from the forward and backward GRU networks, which are presented in the equation (\ref{eq3}). 

\begin{equation}
\mathnormal{r_{ij} = [\overrightarrow{p_{n}}; \overleftarrow{p_{0}}]}
\label{eq3}
\end{equation}

To inject this relation information into the concept representation, we follow the idea of relative position embedding \cite{shaw2018self,salton2017attentive}, which introduces the attention score method based on both the concept representations and their relation representation. To compute the attention score, we split the relation vector $\mathnormal{r_{ij}}$ passed from the linear layer into forward relation encoding $\mathnormal{r_{i\rightarrow j}}$ and backward relation encoding $\mathnormal{r_{j\rightarrow i}}$, as follows:

\begin{equation}
\mathnormal{[\mathnormal{r_{i\rightarrow j}}; \mathnormal{r_{j\rightarrow i}}]} = \mathnormal{W_{r}r_{ij}}
\end{equation}
where $\mathnormal{W_{r}}$ is the parameter matrix. This split renders the model consider bidirectionality of the path.

Thereafter, we compute the attention score considering the concepts and their relations. Note that $\mathnormal{c_i}$ and $\mathnormal{c_j}$ are the concept embedding. The equation is presented below:

\begin{equation}
\begin{aligned}
s_{ij} & = f(c_i, c_j, r_{ij}) \\ 
& = (c_i + r_{i\rightarrow j})W_q^{\top}W_k(c_j + r_{j\rightarrow i} ) \\
& = c_{i}W_q^{\top}W_{k}c_{j} + c_{i}W_{q}^{\top}W_{k}r_{j\rightarrow i} \\
&    + r_{i\rightarrow j}W_{q}^{\top}W_{k}c_{j} + r_{i\rightarrow j}W_{q}^{\top}W_{k}r_{j\rightarrow i}\\
\end{aligned}
\label{eq5}
\end{equation}

The first term in the last line of equation (\ref{eq5}) is the original term in the vanilla attention mechanism, which includes the pure contents of the concept. The second and third terms capture the relation bias with respect to the source and target, respectively. The final term represents the universal relation bias. As a result, the computed attention score updates the concept embedding while maintaining fully-connected communication \cite{cai2019graph:2019}. Therefore, concept--relation interactions can be injected into the concept node vector. The resulted concept representations are summed into the whole graph vector and fed into the Transformer Layers to model the interaction between AMR and ConceptNet concept representation. 
The major advantage of this relation-enhanced attention mechanism is that it provides a fully connected view of input graphs by making use of the relation multi-head attention mechanisms. Since we integrate two different concept types from the AMR graph and ConceptNet into a single graph, the model globally recognizes which path has high relevance to the question during the interpretation.
After obtaining the language and graph vectors, the model concatenates the two vectors, feed these into the Softmax layer, and selects the correct answer.

\section{Experiments}

To show the effectiveness of representing a question using the proposed ACP graph in commonsense reasoning, we conduct four different experiments. 
We first compare the ACP with the ACF graph, which is expanded on all the concepts of the AMR graph, and with the graph that only utilizes the ConceptNet. In addition, we apply our model to three language models that have different encoder structures, showing performance enhancement as shown in Table \ref{table:exp2}. Moreover, we investigate the efficacy of the proposed method on the extended versions of the BERT-base model such as the BERT-large-cased or post-trained BERT model with OMCS data. Finally, we show the performance of our model with official test set.

\subsection{Data and Experimental Setup} 

The \texttt{CommonsenseQA} dataset consists of 12,102
(v1.11) natural language questions and each question has five candidate answers provided by Talmor et al \shortcite{talmor2019commonsenseqa}. As the prediction on the official test set can be evaluated only through the organizers by two weeks, we divide the official training set for the experiment efficiency. We split the given training set into the new training and test sets. The new training, development, and test sets included 8,500, 1,221, and 1,241 examples, respectively. We use RTX8000 for training our model. The parameters for the graph path learning model are identical to the work in Cai \shortcite{cai2019graph:2019}'s model.

\begin{figure}[t]
\begin{subfigure}[b]{.56\linewidth}
\centering
\includegraphics[width=\linewidth]{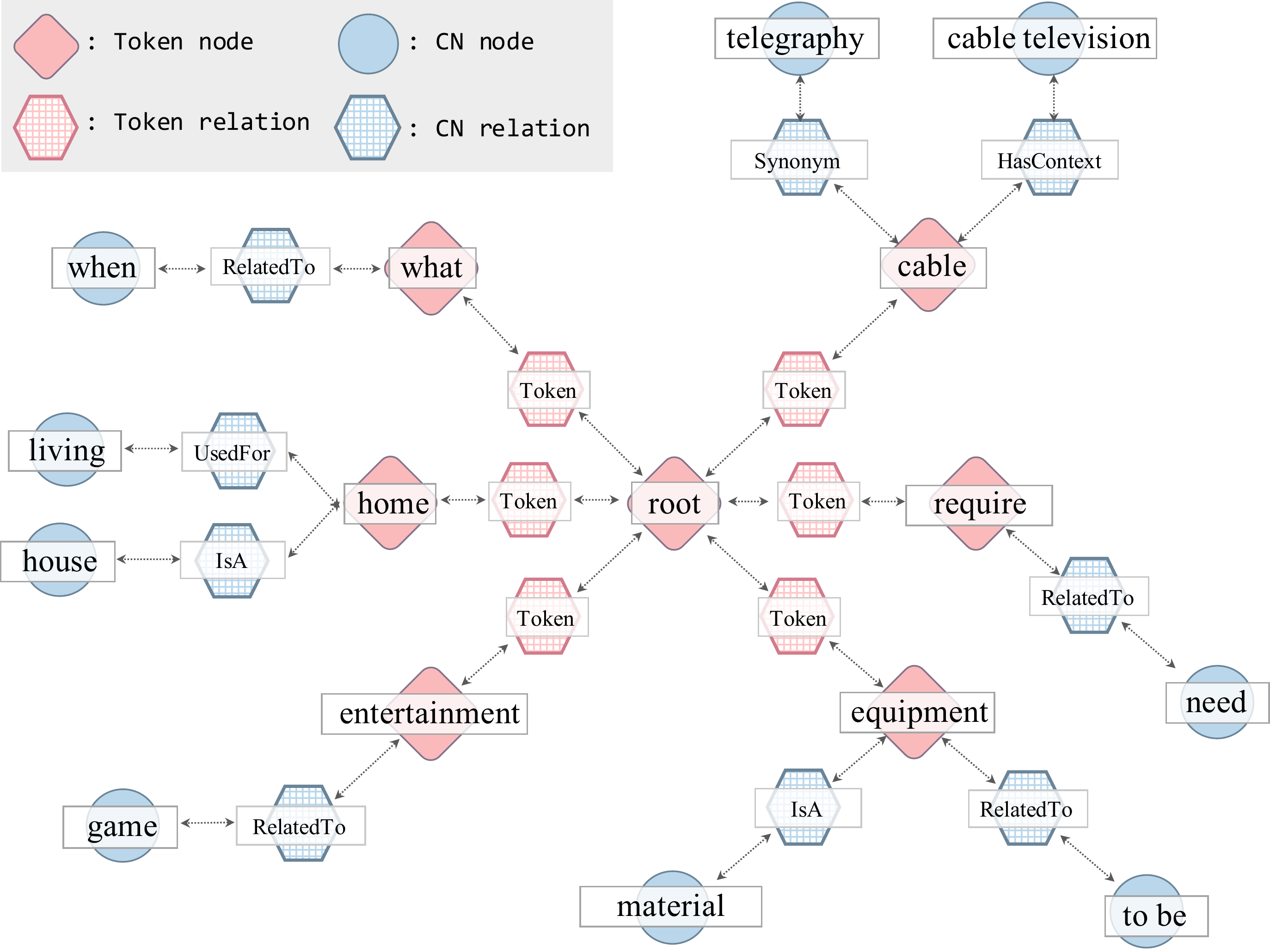}
\caption{ConceptNet -- full}
\label{fig:cn_a}
\end{subfigure}\hfill
\begin{subfigure}[b]{.42\linewidth}
\centering
\includegraphics[width=\linewidth]{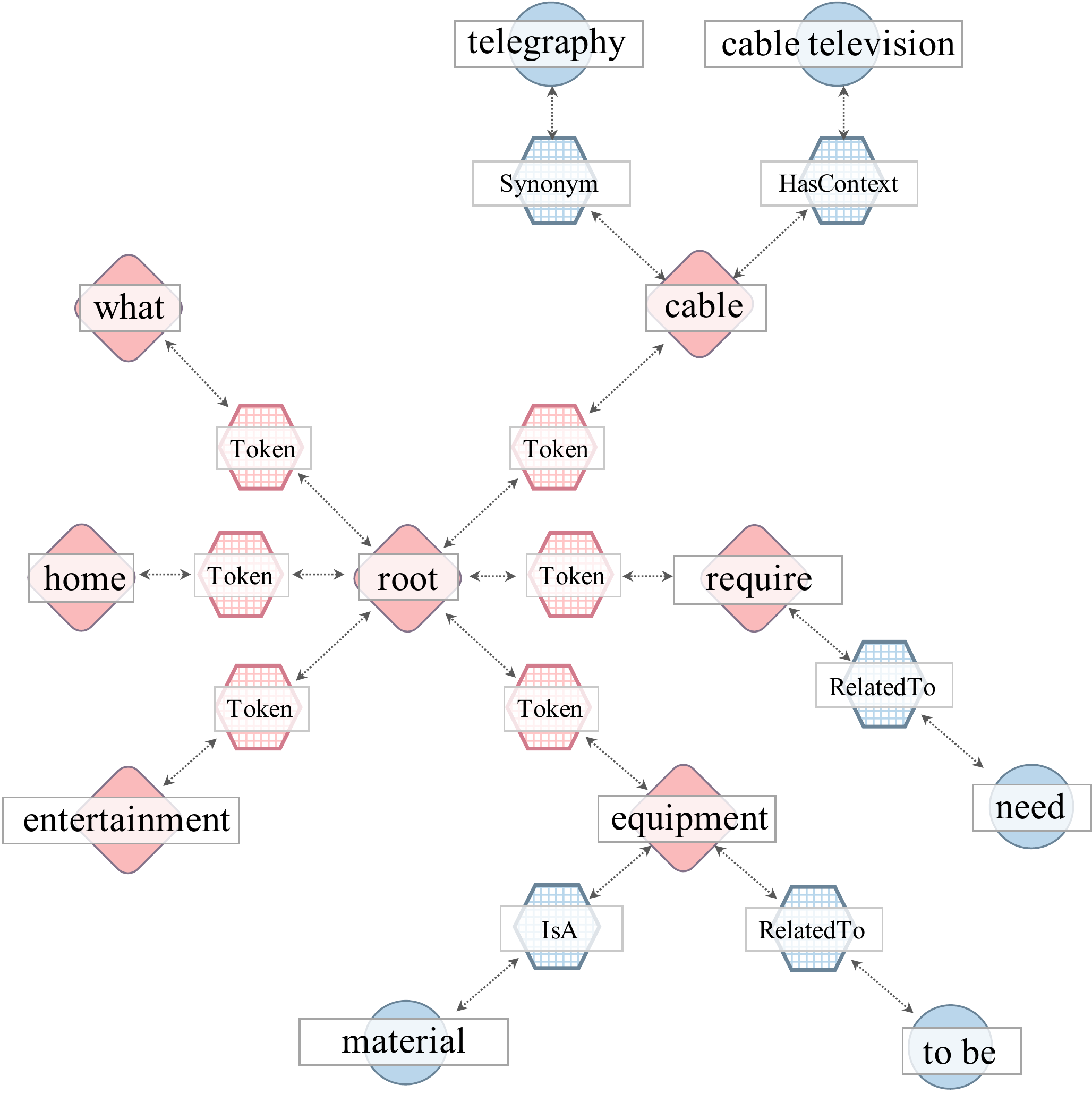}
\caption{ConceptNet -- pruned}
\label{fig:cn_b}
\end{subfigure}%
\caption{Two methods of ConceptNet expansion matched with tokens. The token nodes are all the words in the question}
\label{fig:cn}
\end{figure}

\subsection{Experimental Results}
\textbf{Commonsense reasoning.} 
To demonstrate that our ACP graph is more effective than other graph features, we conduct experiments on diverse graph features that include not only the ACP graph but also the ACF graph. The ACF graph is an integrated graph, with the AMR graph and the ConceptNet matched with all concepts from the AMR graph.
Furthermore, we run experiments only using ConceptNet (CN) in two manners. The one is a method that uses the ConceptNet graph that corresponds to all the question tokens separated by the spaces of the sentence as depicted in Figure \ref{fig:cn} \subref{fig:cn_a}. As the tokens from the question are not connected initially, it may prevent our graph path learning module from reasoning over the CN graph due to the disconnection between the concept nodes. Therefore, we connect all of the tokens from the question to the \texttt{root} node to let our model perform effectively on commonsense reasoning. The other is a method that employed the pruned ConceptNet graph using the the logic of AMR graph as decribed in Figure \ref{fig:cn} \subref{fig:cn_b}. The pruned ConceptNet graph includes the subgraph of the ConceptNet matched with tokens that are connected to \texttt{ARG0} and \texttt{ARG1} of the AMR. For example, as the concept nodes \texttt{require}, \texttt{equipment}, and \texttt{cable} in Figure \ref{fig:amr_cn} \subref{fig:amr_cn_a} are connected with the relation of \texttt{ARG0} and \texttt{ARG1}, only those concept nodes have the ConceptNet relations. Similar to the first method, the tokens from the question are linked with \texttt{token} relation to the \texttt{root} node. 
Note that these two methods do not explicitly make use of the AMR graph concepts.

ACF graph is depicted in Figure \ref{fig:amr_cn} \subref{fig:amr_cn_a} and is expressed as follows.
\begin{equation}
    \mathcal{G_\mathnormal{ACF}}=(\{\mathcal{V_{\mathnormal{amr}}} \cup {\mathcal{V}_\mathnormal{cn}^\mathit{\tiny amr}}\},\{\mathcal{E_{\mathnormal{amr}}} \cup {\mathcal{E}_\mathnormal{cn}^\mathit{\tiny amr}}\})
\label{eq1}
\end{equation}

Note that the AMR graph is denoted as  $\mathcal{G_\mathnormal{AMR}}= \{\mathcal{V_{\mathnormal{amr}}},\mathcal{E_{\mathnormal{amr}}}\}$ and the subgraph of ConceptNet matched with the ACF graph is denoted as $\mathcal{G_\mathnormal{CN}^\mathnormal{AMR}}= \{\mathcal{V_{\mathnormal{cn}}^\mathnormal{amr}},\mathcal{E_{\mathnormal{cn}}^\mathnormal{amr}}\}$.

The (1) CN full graph (CF) and (2) CN pruned graph (CP) are illustrated in Figure \ref{fig:cn} and defined as follows, respectively:

\begin{equation}
    \mathcal{G_\mathnormal{CF}}=(\{ \mathcal{V_{\mathnormal{token}}} \cup \mathcal{V_{\mathnormal{cn}}^\mathnormal{token}}\} \cup \{\mathtt{\texttt{root}}\}\}, \{\mathcal{E_{\mathnormal{cn}}^\mathnormal{token}} \cup  \{\mathtt{\texttt{token}}\}\})
\label{eq1}
\end{equation}
\begin{equation}
    \mathcal{G_\mathnormal{CP}}=(\{\mathcal{V_{\mathnormal{token}}} \cup \mathcal{V}_\mathnormal{cn}^\mathit{\tiny amr^{arg}}\} \cup \{\mathtt{\texttt{root}}\}\} ,\{{\mathcal{E}_\mathnormal{cn}^\mathit{\tiny amr^{arg}}}\cup  \{\mathtt{\texttt{token}}\}\})
\label{eq2}
\end{equation}
\vspace{3mm}


The ConceptNet graph is denoted as  $\mathcal{G_\mathnormal{CN}}= \{\mathcal{V_{\mathnormal{cn}}},\mathcal{E_{\mathnormal{cn}}}\}$ and the subgraph of ConceptNet matched with the question token is denoted as $\mathcal{G_\mathnormal{CN}^\mathnormal{token}}= \{\mathcal{V_{\mathnormal{cn}}^\mathnormal{token}},\mathcal{E_{\mathnormal{cn}}^\mathnormal{token}}\}$. For the CN pruned graph, the notation is as follows: the subgraph of ConceptNet matched with the tokens connected to \texttt{ARG0} and \texttt{ARG1} is defined as $\mathcal{G_\mathnormal{CN}^\mathnormal{AMR^{arg}}}= \{\mathcal{V_{\mathnormal{cn}}^\mathnormal{amr^{arg}}},\mathcal{E_{\mathnormal{cn}}^\mathnormal{amr^{arg}}}\}$. In addition, $\mathcal{V_{\mathnormal{token}}}$ denotes the token of the sentence that is separated by a space. 
\vspace{-5mm}

\begin{center}
\begin{table}[h]
{
\centering
\scalebox{0.80}{
	
	\begin{tabular}{c|lccc}
    		\toprule
    		\textbf{Language Encoder} & \textbf{Graph type} & {Ndev-Acc.(\%)}  & {Ntest-Acc.(\%)}  \\
    		\midrule\midrule
    		\multirow{7}{*}{BERT-base-cased}
    		& - & 51.81 & 51.59 & \\
    		\cmidrule{2-4}
    	    &  AMR - original & 52.82 & 52.78 \\
    		\cmidrule{2-4}
    		& CN -- full (\textit{CF}) & 53.80 & 53.10 \\
    		& CN -- pruned (\textit{CP}) & 52.61 & 52.53 &  \\
            \cmidrule{2-4}
    		& AMR-CN -- full (\textit{ACF}) & 52.98 &  52.94 \\
    	    & AMR-CN -- pruned (\textit{ACP}) & \textbf{53.97} & \textbf{53.58}  \\
    		\bottomrule
	    \end{tabular}
	    }
	\caption{Experiments on diverse graph types. \vspace{12pt}}
\label{table:exp1}
}
\end{table}
\end{center}

As indicated in Table \ref{table:exp1}, whereas the BERT fine-tuning only score 51.59\% in terms of accuracy, the models with the AMR graph or ConceptNet exceed this result, scoring over 52\%. 
Interestingly, the ACP graph achieves the best score among all other graph types. These results demonstrate that the ACF graph and other CF, CP graphs consider all possible paths which are unnecessary and obtain insufficient performance. In other words, the ACP graph enables the Graph Path Learning Module to find reasonable paths efficiently by ignoring the irrelevant paths. Since the ACP graph provides the best results, other experiments are conducted with the ACP graph feature. 

\vspace{3mm}

\begin{flushleft}
\textbf{Extensions of BERT.} We also demonstrate the effects of the proposed method on improved BERT models that are post-trained BERT-base-cased with OMCS and BERT-large. Previous studies mostly addressed the \texttt{CommonsenseQA} utilizing a post-training approach with OMCS data, which is a freely available crowd-sourced knowledge base of natural language statements regarding the world. Because of the high performance on the ACP graph feature in Table \ref{table:exp1}, we illustrate the effect on the performance of the ACP graph with respect to the different version of BERT. As indicated in Table \ref{table:exp3}, our method combined with the post-trained BERT on OMCS achieved 54.31\% in the new test set. Moreover, the BERT-large model with our method outperformed the BERT-large fine-tuning model, obtained 58.98\% in the new test set. 
\end{flushleft}

\begin{center}
\begin{table}[h]
{
\centering

\scalebox{0.85}{
	
    \begin{tabular}{l|cccc}
			\toprule   
			 \textbf{Language model} & {Ndev-Acc.(\%)}  & {Ntest-Acc.(\%)}  \\
			\midrule\midrule
			BERT post-trained w/ OMCS & 52.13 & 52.08\\
			BERT-large-cased  & 57.16 & 56.24 &\\
			\midrule\midrule
			BERT post-trained  w/ OMCS \& AMR-CN -- pruned(\textit{ACP})  & \textbf{54.79} & \textbf{54.31}\\
			BERT-large-cased w/ AMR-CN -- pruned(\textit{ACP}) & \textbf{59.37} & \textbf{58.98} \\ 
			\bottomrule
		\end{tabular}
		}
		\caption{Experiments on other BERT models.}
\label{table:exp3}

}
\end{table}
\end{center}
\vspace{-10mm}
\begin{flushleft}
\textbf{Comparison on different language models.} Table \ref{table:exp2} presents the results of the comparison experiments on different language models with varying transformer encoder structures. Input of the language model is formalized as ``\texttt{[CLS]+Question+[SEP]+candidate answer}.'' All of the language models that use our method outperformed their own fine-tuned score, achieving 53.58\% with BERT-base, 60.35\% with XLNet-base \cite{yang2019xlnet}, 51.08\% with ALBERT-base \cite{lan2019albert}, and 70.91\% with ELECTRA-base \cite{clark2020electra} on our new test set. This implies that the concept representations obtained from the our ACP graph had significant effects and stable generality on \texttt{CommonsenseQA}, regardless of the language model encoder types. 
\end{flushleft}

\begin{center}    
\begin{table}[h]
\centering
\scalebox{0.85}{
	
    \begin{tabular}{l|cccc}
	        \toprule   
			 \textbf{Language Encoder} & {Ndev-Acc.(\%)}  & {Ntest-Acc.(\%)}  \\
			\midrule\midrule
			BERT-base-cased  & 51.81 & 51.59 & \\
		    XLNet-base-cased & 57.98 & 57.05  \\
			ALBERT-base & 50.12 & 49.22   \\
			ELECTRA-base & 71.25 & 70.19 \\

			\midrule\midrule
			BERT-base-cased w/ AMR-CN -- pruned (\textit{ACP}) & \textbf{53.97} & \textbf{53.58} \\ 
			XLNet-base-cased w/ AMR-CN -- pruned (\textit{ACP}) & \textbf{61.01} & \textbf{60.35}\\
			ALBERT-base w/ AMR-CN -- pruned (\textit{ACP}) & \textbf{51.51} &  \textbf{51.08} \\
			ELECTRA-base w/ AMR-CN -- pruned (\textit{ACP}) & \textbf{71.99} &  \textbf{70.91} \\
			\bottomrule
		\end{tabular}
		}
		\caption{Experiments on different language models. }
\label{table:exp2}
\end{table}
\end{center}

\begin{figure}[h]
\begin{subfigure}[b]{.60\linewidth}
\centering
\includegraphics[width=\linewidth]{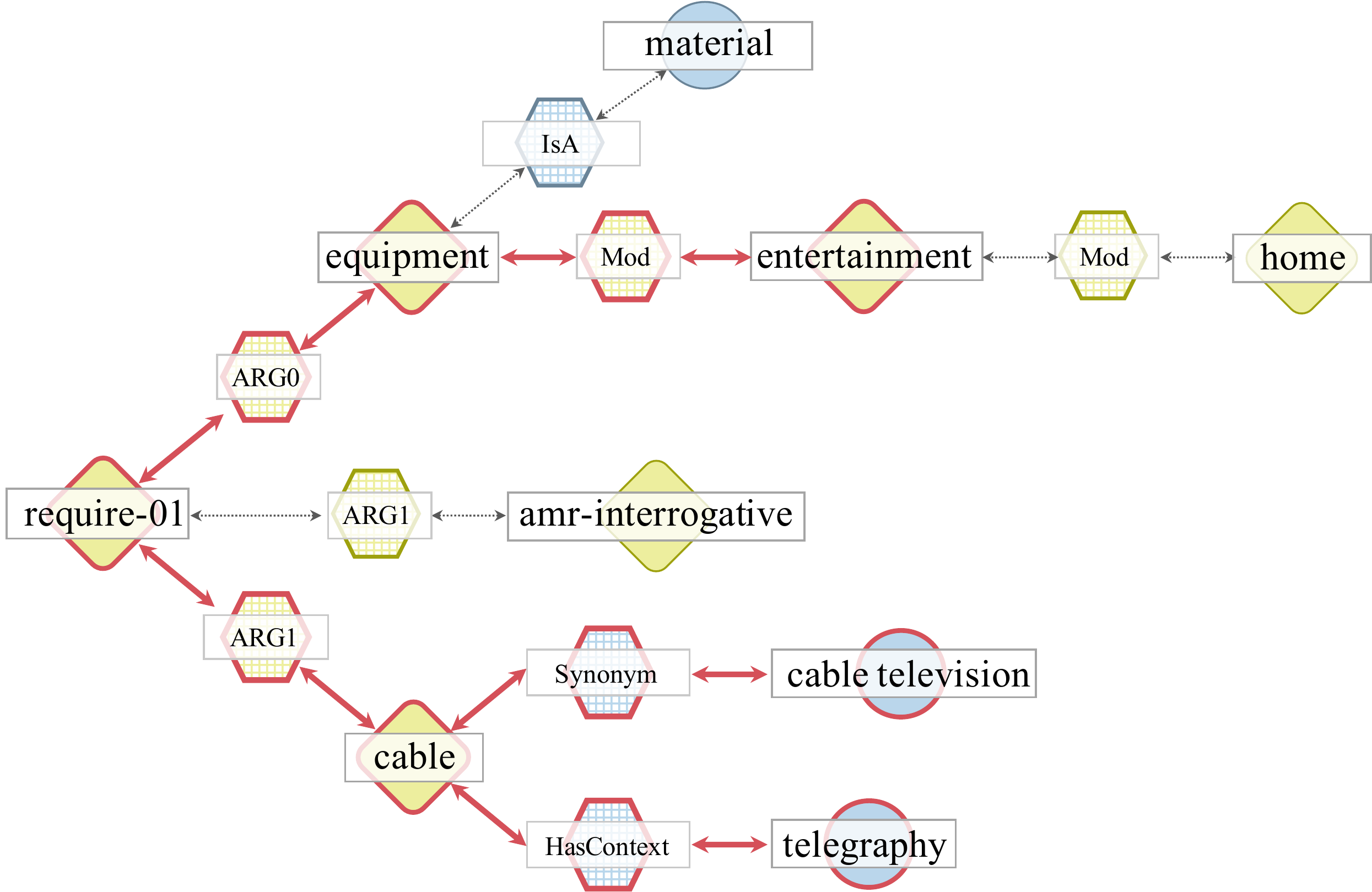}
\caption{AMR-CN extended graph with high attention weight relations}
\label{subfig:case-amrcn}
\end{subfigure}\hfill
\begin{subfigure}[b]{.39\linewidth}
\includegraphics[width=\linewidth]{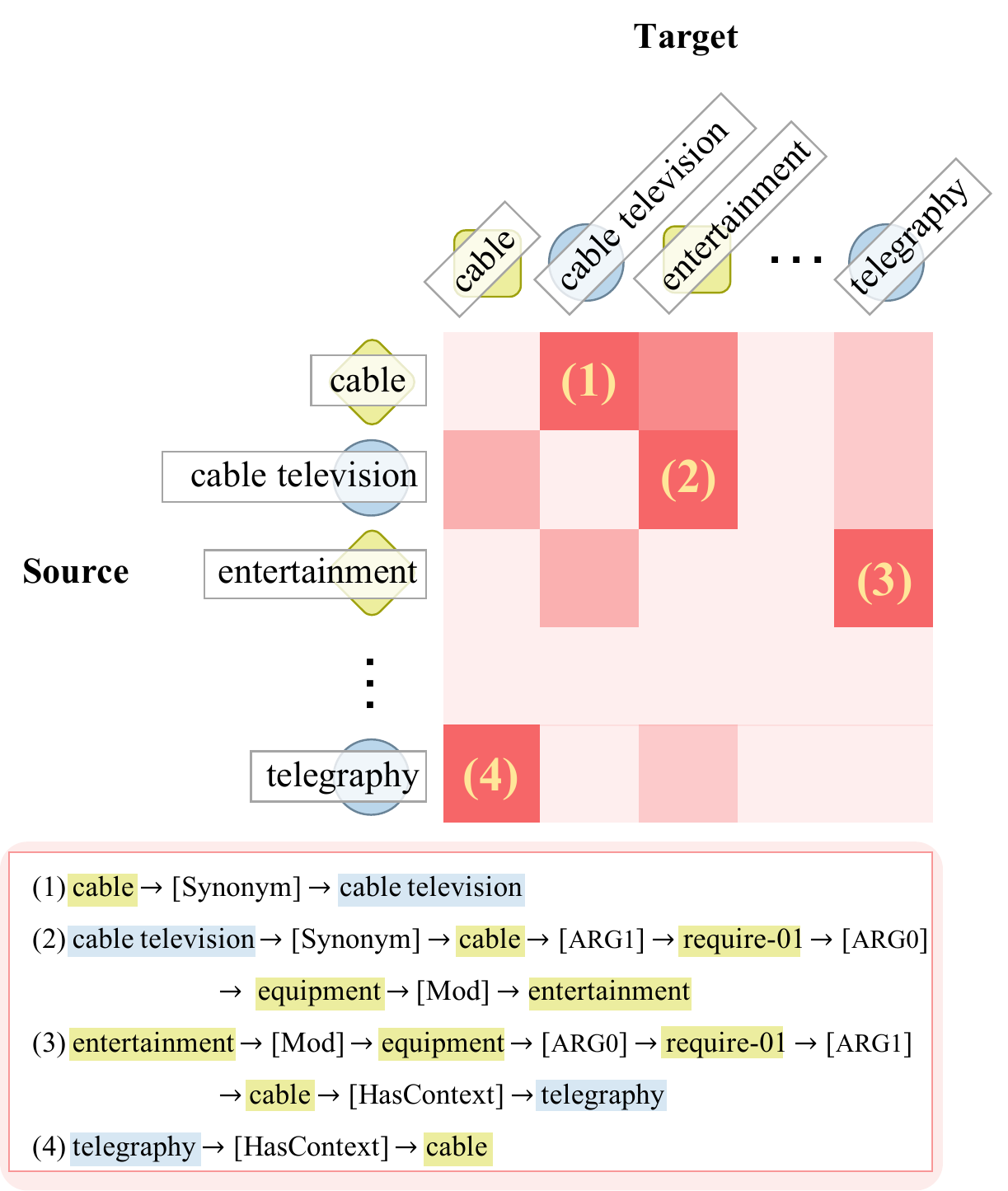}
\caption{Heatmap of path attention from source to target words. The details of each path are presented below the heatmap}
\label{subfig:case-heatmap}
\end{subfigure}%
\caption{Case study of the question ``\texttt{What home entertainment equipment requires cable?}'' and candidate answers \texttt{(a)radio shack}, \texttt{(b)substation}, \texttt{(c)cabinet}, \texttt{(d)television}, \texttt{(e)desk}. Figure \ref{fig:case-amrcn} \subref{subfig:case-amrcn} presents the entire AMR-CN graph marked with the red color for high attention weight paths in BERT-base-cased w/ AMR-CN--pruned graph (ACP). Figure \ref{fig:case-amrcn} \subref{subfig:case-heatmap} displays the heatmap of the attention weights with respect to the source and target tokens in the path. The details of the paths are also provided.}
\label{fig:case-amrcn}
\end{figure}

\begin{flushleft}
\textbf{Experiment on official test set} As the ELECTRA-base model with the ACP graph shows the highest performance on the new test set, we conduct the experiments on the official test set (1140 examples) with the official training set utilizing ELECTRA-large. For official test set, our model achieve 75.43\% accuracy. 
\end{flushleft}

\begin{center}
\begin{table}[h]
\centering
\scalebox{0.85}{
	
    \begin{tabular}{l|cccc}
			\toprule   
			\textbf{Models} & {Odev-Acc.(\%)}  & {Otest-Acc.(\%)}  \\
			\midrule\midrule
			KagNet \cite{lin2019kagnet}\textsuperscript{$\dagger$} & 64.46 & 58.90 \\
			HyKAS \cite{ma2019towards}\textsuperscript{$\dagger$}  & 80.10 & 73.20 \\
			XLNet + Graph Reasoning \cite{lv2019graph}\textsuperscript{$\dagger$} & 79.30 & 75.30 \\
			ELECTRA-large w/ AMR-CN -- pruned (\textit{ACP}) & \textbf{82.15} & \textbf{75.43} \\
			\bottomrule
		\end{tabular}
		}
		\caption{Experiment with ELECTRA with full training set on the official test set. $\dagger$ denotes the results of the models that use ConceptNet, taken from the official commonsenseQA leaderboard\footnotetext{https://www.tau-nlp.org/csqa-leaderboard} in April 2020.} 
		
\label{table:exp4}
\end{table}
\end{center}

\section{Discussion}
\subsection{Error Analysis}
In some cases of failure, our model exhibits two problems as follows:
\begin{itemize}
\item \textbf{Difficulty in discriminating hard distractors}: All candidate answers from the \texttt{CommonsenseQA} possess a hard distractor, which shares the same relation with the question. When the hard distractor exists in the ACP graph in the path learning module, it also uses the paths of the distractor, instead of the those of correct answer. This may make the model confused as it considers the distractor as the correct answer.

\item \textbf{AMR graph generation error}: Since the AMR graph is generated from the pre-trained model, our model is at risk of using an incorrect AMR graph. An incorrectly produced AMR graph may lead the model to incorrect interpretation and distortion in the wrong direction during the path reasoning process. For example, the AMR graph generated from the question ``\texttt{What can help you with an illness?}'' is described below. 
\begin{verbatim}
(vv1 / possible
      :ARG1 (vv2 / help-01
            :ARG0 (vv3 / amr-unknown)))
\end{verbatim}
As the concept node \texttt{illness} disappeared while generating the graph, our model may not have enough information for extracting the subgraph from ConceptNet. 
\vspace{3mm}
\end{itemize}

\subsection{Case Study} 

The red edges in Figure \ref{fig:case-amrcn} present the paths that have high attention weight for the question ``\texttt{What home entertainment equipment requires cable?}'' In Figure \ref{fig:case-amrcn} \subref{subfig:case-heatmap}, the top four paths with high attention weights are described. As opposed to predicting the answers simply with the ConceptNet graph connected to the question, we allow our model to learn relevant paths inherent in the ACP graph. That is, our graph path learning module with ACP graph is capable of commonsense reasoning exploring the paths.

\vspace{5mm}
\section{Conclusions and Future Works}
We introduce a new commonsense reasoning method, using the proposed ACP graph. This method outperformed the model that simply learns the ConceptNet graph. Furthermore, our method can explain the answer-inference process by interpreting the logical structure of the sentences within commonsense reasoning process. Models that applied our method exhibit higher performance compared to the  previous models. However, certain problems still remain. Though the relations \texttt{ARG0} and \texttt{ARG1} occupy most of the core roles in the AMR graph, it is still arguable that the other choice of relations may lead to better results. Therefore, we will show the experimental results according to the different pruning rules on the \texttt{CommonsenseQA} task in the future. Also, we plan to develop an end-to-end learning model that incorporates the AMR generation model and the question-answering model to reduce the error propagation from the AMR generation.

\section{Acknowledgement}
This work was supported by Institute for Information \& communications Technology Planning \& Evaluation(IITP) grant funded by the Korea government(MSIT) (No. 2020-0-00368, A Neural-Symbolic Model for Knowledge Acquisition and Inference Techniques). Also, this research was supported by the MSIT(Ministry of Science and ICT), Korea, under the ITRC(Information Technology Research Center) support program(IITP-2020-2018-0-01405) supervised by the IITP(Institute for Information \& Communications Technology Planning \& Evaluation)



\bibliographystyle{coling}
\bibliography{coling2020}
\newpage

\begin{flushleft}
\textbf{Appendix A. Related Works}
\vskip 0.2in
We briefly describe the structures and properties of ConceptNet, which consists of commonsense relations. Furthermore, we review previous studies and AMR, which is an essential part of our model.
\end{flushleft}
\vspace{-2mm}
\textbf{ConceptNet.} In ConceptNet \cite{speer2017conceptnet}, real-world assertions are represented as two nodes and directed edges, which denote certain concepts and their relations, respectively. The nodes represent words or phrases from natural language sentences. The edges represent the relations between nodes, and they contain lexical as well as commonsense relation information. As ConceptNet is created by collecting data from various types of knowledge bases, nodes of different types also exist. Each node represents a slightly different meaning considering its role in the sentence. For example, the word ``\texttt{person}'' can be found in the concept of ``\texttt{person/n},'' which is analyzed as a noun with a POS tagger, and with more detailed semantic information, it can be identified as ``\texttt{person/n/wn/body}.'' This information makes possible the detailed extraction of knowledge that considers the purpose of each sentence. Meanwhile, one or more edges may be defined between two nodes. For example, the edge between the nodes ``\texttt{person}'' and ``\texttt{eat}'' can be defined independently as ``\texttt{CapableOf}'' and ``\texttt{Desires}.'' Various concepts and their relations are defined as nodes and edges in ConceptNet, considering the ambiguity in the sentences.

\textbf{Commonsense reasoning.}  Commonsense reasoning is the process of logical inference by using commonsense information. In \texttt{CommonsenseQA}\footnote[1]{https://www.tau-nlp.org/csqa-leaderboard} task, the fine-tuning approach with pre-trained language representations makes use of external commonsense knowledge. There are two means of exploiting external knowledge. The first\footnote[7]{ttps://drive.google.com/file/d/1sGJBV38aG706EAR75F7LYwCqci9ocG9i/view,\\ https://gist.github.com/commonsensepretraining/507aefddcd00f891c83ebf6936df15e8} is the method that post-trained with some commonsense sentence corpus. It then performs fine-tuning with evidence derived from questions and answers. The second method \cite{lv2019graph,lin2019kagnet} is to encode commonsense knowledge graphs and train with language models. The language models that have exhibited high performance in this method are BERT \cite{devlin2018bert}, RoBERTa \cite{liu2019roberta}, which use bidirectional transformer encoders. They also include XLNet \cite{yang2019xlnet}, which is based on autoregressive language modeling, ALBERT \cite{lan2019albert}, which adopts cross-layer parameter sharing and factorized embedding parameterization and ELECTRA \cite{clark2020electra} that is pre-trained with Replaced Token Detection (RTD) task.

\textbf{AMR.} AMR \cite{banarescu2013abstract} represents the relations between concept nodes using the PropBank frameset and vocabularies from the sentences. The edges between two or more concept nodes or the argument nodes are relations. AMR represents semantic roles such as core and numbered roles, and uses more than 100 semantic relations, including negation, conjunction, command, and wikification. In PropBank \cite{bonial2012english}, the semantic roles are labeled in the form of \texttt{ARG0}$\sim$\texttt{4} and \texttt{ARGM}. In general, \texttt{ARG0} denotes the agent of the verb, \texttt{ARG1} is the patient, \texttt{ARG2} means the instrument, benefactive, or attribute, \texttt{ARG3} is interpreted as the starting point, benefactive, or attribute, and \texttt{ARG4} represents the ending point. The root node serves as the central point of the representation and is called frame node. Thereafter, other concept nodes are sequentially combined according to the semantic relations. AMR consists of concept nodes in a single graph that is traversable to all nodes, similar to a parse tree. However, unlike the parse tree, which represents the explicit structure of sentences, AMR aims to describe the conceptual and semantic structure. That is, if the semantic meanings of explicitly different sentences are the same, they can be represented by the same AMR graph. For example, the two sentences ``\texttt{The boy is a hard worker}'' and ``\texttt{The boy works hard}'' are represented by the same PENMAN graph, namely \texttt{(w / work-01 :ARG0 (b / boy) :manner (h / hard))}. The data constructed to generate and evaluate these representations are AMR 2.0 (LDC2017T10) and AMR 1.0 (LDC2014T12). The model with the highest performance on these data was presented by Zhang et al. \shortcite{zhang2019amr:2019}, using BERT. Various NLP fields have exploited AMR, such as sentence generation \cite{cai2019graph:2019,guo2019densely}, summarization \cite{vlachos2018guided,liao2018abstract}, question and answering \cite{mitra2016addressing}, dialogue systems \cite{bonial2019augmenting}, paraphrase detection \cite{issa2018abstract}, and biomedical text mining \cite{wang2017dependency,garg2016extracting}.

\end{document}